\documentclass[a4paper]{article}

\usepackage{INTERSPEECH2016}

\usepackage{graphicx}
\usepackage{amssymb,amsmath,bm}
\usepackage{textcomp}
\usepackage{color}

\usepackage{algorithm}
\usepackage{algorithmic}
\newcommand{\argmax}{\mathop{\rm arg~max}\limits}
\newcommand{\argmin}{\mathop{\rm arg~min}\limits}
\ninept

\title{Automatic Pronunciation Generation by Utilizing a Semi-supervised Deep Neural Networks}

\makeatletter
\def\name#1{\gdef\@name{#1\\}}
\makeatother \name{{\em Naoya Takahashi$^1$, Tofigh Naghibi$^2$, Beat Pfister$^3$}}

\address{$^1$Sony Corporation, Japan \\
  $^2{}^,{}^3$Speech Processing Group, ETH Zurich, Switzerland \\
  {\small \tt NaoyaA.Takahashi@jp.sony.com, \{naghibi, pfister\}@tik.ee.ethz.ch}
}

\begin{document}

  \maketitle
  \begin{abstract}
    Phonemic or phonetic sub-word units are the most commonly used atomic elements to represent speech signals in modern ASRs.
    However they are not the optimal choice due to several reasons such as: large amount of effort required to handcraft a pronunciation dictionary, pronunciation variations, human mistakes and under-resourced dialects and languages.  Here, we propose a data-driven pronunciation estimation and acoustic modeling method which only takes the orthographic  transcription to jointly estimate a set of sub-word units and a reliable dictionary. Experimental results show that the proposed method
    which is based on semi-supervised training of a deep neural network largely outperforms phoneme based continuous speech recognition on the TIMIT dataset.
    
  \end{abstract}

    \begin{figure*}[t]
        \centering
        \includegraphics[width=\linewidth]{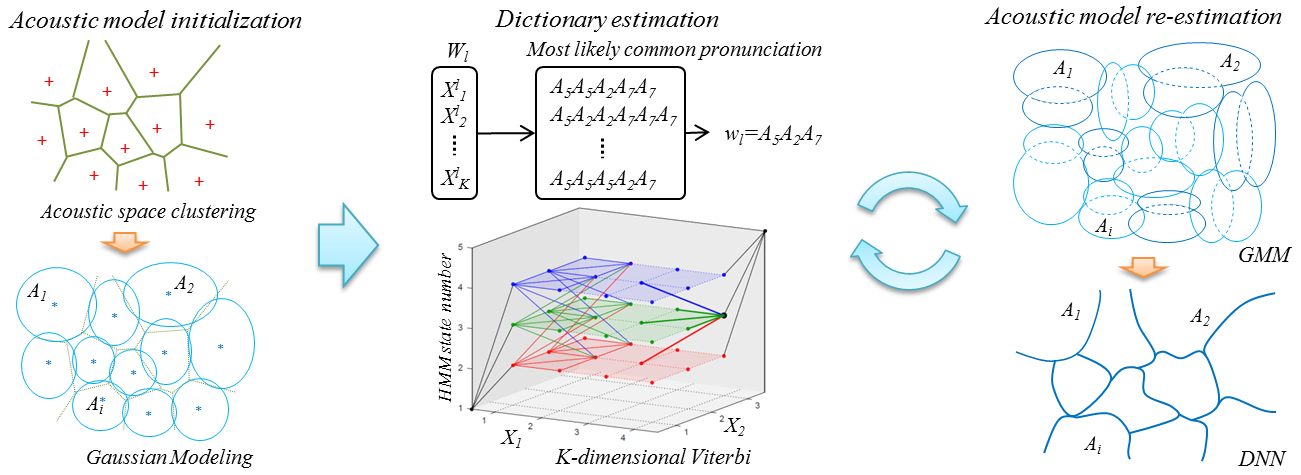}
        \caption{{\it Framework of joint sub-word and dictionary learning. K-dimensional Viterbi illustrated in case of $K=2$.}}
        \label{fig:overview}
    \end{figure*}

  \section{Introduction}
    The three principal resources typically required for developing a phoneme based automatic speech recognizer (ASR) are: 
	transcribed acoustic data for acoustic model estimation, text data for language model estimation, and a pronunciation dictionary to map words to sequences of sub-word units.
	Manual preparation of such resources requires significant investment and expertise.
    Therefore, an automatic generation of pronunciation dictionary from the data is clearly required for many dialects and languages.
	
	Developing ASRs for dialects and under-resourced languages has attracted growing attention over the past few years \cite{Das2015,Qian2011, Besacier2014}.  A main challenge to develop ASR for under-resourced domains is to produce a reliable pronunciation dictionary from limited available resources.  For major languages, however, a canonical pronunciation dictionary is usually already available. However, such dictionaries may be error-prone due to the fact that they are manually generated and in most cases do not cover pronunciation variants. There were several attempts to tackle these problems \cite{Saraclar2000,Wester2003, Hain2005, Mcgraw2013}.
	
	Lu et al. \cite{Singh2002} proposed a data-driven dictionary generator to include new pronunciations based on newly coming acoustic evidence.
	Goel et al. in \cite{Ghoshal2010} use a grapheme-to-phoneme approach to guess the pronunciation and iteratively refine the acoustic model and the dictionary. However, these methods still require a high-quality initial pronunciation dictionary created by an expert.
	
	In modern ASRs words are represented by smaller sub-word units such as phonemes and the pronunciation dictionary maps words to sequences of sub-word units. However, sub-word units do not essentially need to be linguistically motivated elements. 
	In fact, given a set of acoustic samples, the linguistically defined units are most probably not the optimal ones for speech recognition \cite{Naghibi2013}.
	For instance telephony speech, where high frequency components have been filtered out, requires a modified dictionary with slightly different set of fricatives than full-bandwidth speech. 

    Over the past few years, there have been several attempts to move beyond phoneme based sub-word units by jointly learn a set of sub-word units and their corresponding dictionary directly from the given data \cite{Holter1997,Bacchiani1999,Singh2002}.
	Bacchiani and Ostendorf \cite{Bacchiani1999} proposed an iterative acoustic segmentation and clustering approach to build sub-word units from speech signals and subsequently construct the dictionary based on the estimated sub-word units. Singh et al. \cite{Singh2002} introduced a divide-and-conquer strategy to recursively update sub-word units and dictionary. The dictionary computation was done by means of an n-best type algorithm which is known to produce sub-optimal solutions. Although their approach demonstrates some promising results, the performance is still not comparable with a phoneme based ASR.

    
	The main focus of this paper is to design an ASR based on an automatically generated dictionary that outperforms commonly used phoneme based ASRs. While most of the solutions proposed to find a pronunciation based on multiple utterances of a word are n-best type heuristics \cite{Singh2002,Svendsen2004, Mokbel1999}, in this paper, we employ an approximation of the K-dimensional Viterbi algorithm proposed in our previous works \cite{Gerber2011, Naghibi2013}. This approach gives us the maximum-likelihood estimates of the pronunciations. These high-quality  pronunciations are one of the key factors to outperform phoneme based ASRs. Moreover, to learn proper sub-word units, we combine the strength of Gaussian mixture models (GMM) and deep neural network (DNN) based acoustic modeling. We formulate this problem as an instance of a semi-supervised self-learning process. By taking advantage of the robustness of hidden Markov models (HMM) with GMM based observation probability distribution  against labeling errors, we train the first set of sub-word units and output the first set of pronunciations. We then use this dictionary to re-label the data and employ the higher expressiveness of DNNs to improve the modeling of sub-word units and the dictionary in an iterative process. In each iteration round, a new dictionary is generated and by means of this new dictionary the data is re-labeled. This data is again used to train the DNN. As shown in the experiments, the proposed results achieves more than 10\% absolute improvement over the phoneme based approach on TIMIT data in a continuous speech recognition task. 
	
	
    The reminder of this paper is organized as follows. The proposed framework and its components for joint sub-word units and dictionary learning are introduced in Section~\ref{sec:methods}. In Section~\ref{sec:exper} the experimental results are demonstrated and finally, conclusions are summarized in Section \ref{sec:concl}.

  \section{Semi-supervised joint Dictionary and Acoustic Model Learning}
  \label{sec:methods}
    \subsection{Framework}
	In the rest of this paper, we refer to data-driven sub-word units as abstract acoustic elements (AAEs) in contrast to phones. Our goal is to jointly learn the pronunciation dictionary $d^* = \{\omega_1,\cdots,\omega_L\}$ of $L$ pronunciations $\omega_i$ and $N$ AAE models $\lambda^*=\{A_1,\cdots,A_N\}$ that maximize the joint likelihood:
    \begin{equation}
        \lambda^*,d^* = \argmax_{\Lambda,D} P(\textbf{X}|\textbf{T},\Lambda,D)
    \label{eq:origi}
    \end{equation}
    where $\textbf{X}=(X_1, \cdots,X_M)$ is the set of training utterances, $\textbf{T}=(T_1, \cdots,T_M)$ is the set of corresponding orthographic transcriptions, $M$ is the number of utterances, $\Lambda$ is the universe of all possible sets of $N$ AAEs and $D$ is the universe of all the dictionaries which map words to AAEs sequences.
    It is hard to find the optimal solution for the optimization problem in (\ref{eq:origi}) due to its complex non-linear nature. It is thus decomposed into two simpler optimization problems which can be solved iteratively. 
    \begin{eqnarray}
        d^i = \argmax_{D} P(X|T,\lambda^i,D) \label{eq:dicup}\\
        \lambda^{i+1} = \argmax_{\Lambda} P(X|T,\Lambda,d^i) \label{eq:acup}
    \end{eqnarray}
    Since the pronunciation of each word can be estimated independently from other words, the dictionary estimation in (\ref{eq:dicup}) can be decomposed into $L$ maximum likelihood estimations as follows:
    \begin{equation}
        \begin{split}
        \omega_l = &\argmax_{\omega} \prod_{j\in \Omega_l} \max_{\textbf{S}_j} P(X_j,\textbf{S}_j|\lambda)\\
        &\text{subject to: }\textbf{S}_j \in \mathbb{S}_\omega
        \end{split}
    \label{eq:jointlh}
    \end{equation}
    where $\Omega_l$ is the set of indices of utterances of word $W_l$, $\textbf{S}_j$ is a sequence of AAEs and $\mathbb{S}_\omega$ denotes a set of all possible AAE sequences of the pronunciation $\omega$. For instance in $\mathbb{S}_\omega$, if the pronunciation is $\omega=A_1A_2A_3$, some samples in $\mathbb{S}_\omega$ may be $A_1A_1A_1A_2A_3, A_1A_2A_2A_3A_3$ and $A_1A_1A_2A_3A_3A_3$. The constraint in (\ref{eq:jointlh}) implies that all AAE sequences should be samples of the same pronunciation.
    For the case where $\lambda$ is modeled by a left-to-right HMM without skips, which is the most common topology in HMM based ASRs, a solution of (\ref{eq:jointlh}) has been proposed in \cite{Gerber2011} (Details are in Section~\ref{sec:dicupdate}.).
    In (\ref{eq:acup}), since the dictionary is fixed, the problem results in a common acoustic model estimation given the dictionary. However, the labels re-assigned by using the estimated dictionary are very noisy since the dictionary is automatically estimated from data without any expert supervision. Therefore, a robust model is required at early stage of the training iteration while a more expressive and powerful model such as a DNN \cite{Hinton2012, Abdel-hamid2012} can be used after the reliable dictionary is obtained. 
    
    The joint dictionary and AAE learning framework is illustrated in Figure \ref{fig:overview} and summarized as follows:
        

    \begin{algorithm}                      
        \caption{ Semi-supervised joint AAEs and dictionary learning}         
        \label{alg1}                          
        \begin{algorithmic}[1]
        \STATE $i=0$ \\
        // Initialize AAE models $\lambda^0$ (Section~\ref{sec:init})
        \STATE Clustering the acoustic space.
        \STATE Model each cluster by GMM and set as $\lambda^0$.\\
        // Start joint AAEs and dictionary learning 
        \WHILE{ ( Performance is improved ) } 
            \STATE Given AAE models $\lambda^i$, update dictionary $d^i$ by maximizing joint likelihood multiple utterances (Section~\ref{sec:dicupdate}).
            \STATE Given dictionary $d^i$, double the number of mixtures and update AAE models $\lambda^{i+1}$ (Section~\ref{sec:acupdate}).
            \STATE $i \gets i + 1$
        \ENDWHILE
        \STATE Replace GMM by DNN and train AAE model using labels obtained by HMM-GMM (Section~\ref{sec:acupdate}).
        \WHILE{ ( Performance is improved ) } 
            \STATE Given AAE models $\lambda^i$, update dictionary $d^i$ by maximizing joint likelihood multiple utterances.
            \STATE Given dictionary $d^i$, re-train DNN based AAE models $\lambda^{i+1}$ (Section~\ref{sec:acupdate}).
            \STATE $i \gets i + 1$
        \ENDWHILE       
        \end{algorithmic}
    \end{algorithm}

    \subsection{Acoustic Model Initialization}
    \label{sec:init}
    Initial AAE models can simply be obtained by clustering the acoustic space. The acoustic space can be described by any feature as long as it is informative enough to discriminate between different words. 
    We employed the Linde-Buzo-Gray (LBG) algorithm \cite{Linde1980} with a squared-error distortion measure to cluster the acoustic feature vectors. 
    The LBG clustering algorithm tends to assign more codebook vectors to high-density areas which is a useful property in order to obtain discriminative initial AAEs.
    Each cluster is then modeled by a GMM with a single Gaussian component. These models are used as the initial models for AAEs. 
    
    \subsection{Dictionary Generation}
    \label{sec:dicupdate}
    The solution of (\ref{eq:jointlh}) proposed in \cite{Gerber2011} is an extension of the standard one-dimensional Viterbi algorithm to $K$ dimensions. The K-dimensional Viterbi algorithm calculates the most probable HMM state sequence which is common to $K$ given utterances.  While this algorithm is rigorous, its complexity grows exponentially with the number of utterances, which consequently makes it infeasible to apply it to more than a few utterances. An efficient approximation of the K-dimensional Viterbi algorithm has been proposed in \cite{Naghibi2013} where the problem to find the joint alignment and the optimal common sequence for $K$ utterances is decomposed into $K{-}1$ applications of two-dimensional Viterbi algorithm. This approximation starts with finding the best alignment between two utterances. Then, while keeping the alignment between the already processed utterances fixed,  the next utterance is aligned with this master utterance. The AAE sequence of the final master utterance is the approximation of the K-dimensional Viterbi pronunciation.

    \subsection{Acoustic Modeling}
    \label{sec:acupdate}
    Once the dictionary is updated, all utterances are decoded based on the new pronunciation of the words in the dictionary and the AAEs are re-estimated according to the new labels.
    The AAEs can be modeled by commonly used models such as HMM/GMM or HMM/DNN. However, at the beginning of the training iteration, the model and dictionary are not accurate enough and more probable to get stuck in a bad local optimum if the model's degree of freedom is too high. In order to avoid this situation, we start the training with a simple model, namely one Gaussian component for each AAE with a diagonal covariance matrix. In each iteration, the dictionary gets more accurate. Thus, the number of mixture components are doubled in order to increase the modeling power. Once the performance is saturated the GMM is replaced with the DNN in order to utilize more expressive modeling capability. This process makes the semi-supervised DNN training feasible and prevents it to get stuck in a bad local optimum.
    The HMM state-level transcription is obtained by force-aligned decoding with optimised HMM-GMM and dictionary. This transcription provides labels for DNN training. The DNN is trained to estimate HMM posterior states by minimizing the cross entropy loss $L$ with $l_1$ regularization using back propagation:
    \begin{eqnarray}
        \argmin_{W} \sum_{i,j} L(\textbf{x}^i_j,y^i_j,W)+\rho\|W\|_1 \label{eq:obj}
    \end{eqnarray}
    where $\textbf{x}^i_j\in X_i$ is the $j$th feature vector of the $i$th utterance, $y_j^i$ is the corresponding label and $W$ is the set of network parameters, respectively. $\rho$ is a constant parameter which is set to $10^{-6}$ in this work.

  \section{Experiments}
  \label{sec:exper}
    We conducted several sets of experiments on the TIMIT corpus \cite{timit}. The TIMIT corpus provides a manually prepared dictionary and phone-level transcriptions with 61 phones. As a baseline, 61 phone models were trained using the TIMIT dictionary and the provided transcriptions.  
    We used 12 mel frequency cepstral coefficients (MFCCs) and energy with their deltas and delta-deltas as descriptors of the acoustic space. The speech data was analyzed using a 25 ms Hamming window with a 10 ms frame shift.
    We evaluated phone based DNN-HMM, GMM-HMM and AAE based GMM-HMM model as baselines. The DNN architecture was comprised of 7 hidden layers. The first hidden layer had 2048 nodes, next 5 layers had 1024 nodes and the number of nodes at the last layer was equal to the number of HMM states to be predicted. All hidden layers were equipped with the Rectified Linear Unit (ReLU) non-linearity \cite{Dahl2013}. The input to the network was 11 contiguous frames of MFCCs. The networks were trained using mini-batch gradient descent based on back propagation with momentum. We applied dropout \cite{Hinton2012} to all hidden layers with dropout probability $0.5$. The batch size was set to 128. HMMs had left-to-right, no-skipping topology with three states for each phoneme as opposed to one state for each AAE. HMMs were trained using a modified version of HTK \cite{htk341} and DNNs were implemented using Lasagne \cite{lasagne}.
    
    \subsection{Isolated Word Recognition}
      The first set of experiments were on the isolated word recognition to test the performance of the proposed methods and investigate the effects of hyper parameters such as the number of mixture components and the number of AAEs.
      For joint pronunciation estimation and acoustic models training, we collected a pronunciation training set comprising of words with more than 10 utterances from the TIMIT training set. The total number of utterances in the pronunciation training set was 12800. 
      After excluding words with less than 4 characters (e.g., a and the), 339 distinct words were collected from the TIMIT test set for the isolated word speech recognition task, resulting in 3900 utterances in total.
      The baseline GMM based phone models were trained with 32 mixture components.
      During the GMM based AAE model training the number of mixtures was doubled for each iteration until it reached 128 mixtures as described in Section~\ref{sec:acupdate}.
      
      \subsubsection{Comparison with phonetic approach} 
      The word error rates (WER) of each method are shown in Table~\ref{tab:iwr}. The results show that the proposed data-driven method clearly outperforms the baseline methods. The proposed AAE-DNN method achieved 10.3\% and 2.4\% improvement over GMM and DNN based phonetic acoustic models, respectively. This suggests that a more accurate dictionary and better acoustic models can be obtained directly from training data without any human expertise. Moreover, AAE-DNN method improves the performance by 3.2\% over the AAE-GMM method. This indicates that the DNN was successfully trained in the semi-supervised manner and the final model could effectively use the its expressive modeling power.
      \begin{table}[t]
        \caption{\label{tab:iwr} {\it Comparison of word error rates of each method on 339 words isolated word recognition (\%). Baseline phone models are trained by using the TIMIT dictionary.}}
        \vspace{2mm}
        \centering{
          \begin{tabular}{ c | r } 
            \hline
            Method	& WER \\
            \hline
            Phone GMM	& 18.18 \\
            Phone DNN	& 10.31 \\
            \hline
            AAE GMM	    & 11.15 \\
            AAE DNN	    & \textbf{7.93} \\
            \hline
          \end{tabular}
        }
      \end{table}
      
      \subsubsection{Number of AAEs}
\begin{figure}[t]
            \centering
            \includegraphics[width=\linewidth]{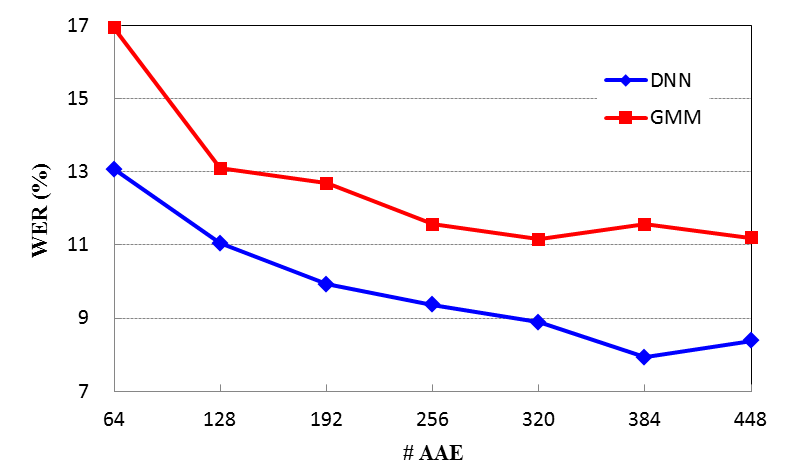}
            \caption{{\it Performance of AAE based recognizers with different number of AAEs on test set with 339 words}.}
            \label{fig:NAAE}
        \end{figure}

      Our second experiment focused on the effects of the number of AAEs, i.e. $N$. We trained the dictionary and AAE models with $N = 64, 128, 192, 256, 320, 384, 448$. The word error rates of DNN and GMM based AAE models are illustrated in Figure \ref{fig:NAAE}. The number of mixtures of the GMMs were determined experimentally as shown in Table~\ref{tab:ngmm}. For DNN based AAE models, the best result are obtained with 384 AAEs in contrast to with 320 AAEs for the GMM based models. Interestingly, the optimal number of AAE states is far higher than the number of states of the phone models (61 phonemes $\times$ 3 states = 183 states). This is an indication that the proposed data-driven approach to jointly generate the sub-word units and dictionary models the acoustic space more precisely than the linguistically motivated phonetic units and the manually designed dictionary. It is also worthwhile to mention that the optimal number of DNN based AAE models was higher than that of GMM based models.
      This is perhaps due to the fact that the DNN was trained discriminatively, allowing to efficiently model the interaction between higher number of AAEs.
        \begin{table}[t]
            \caption{\label{tab:ngmm} {\it Word error rates in \% of AAE based recognizers with different number of AAEs and GMM mixture. The best performance for each number of AAE is plotted in Figure \ref{fig:NAAE}}.}
            \vspace{2mm}
            \centering{
            \begin{tabular}{ c | c c c c} 
                \hline
                \# of AAE & \multicolumn{4}{c}{\# of mixture}\\				
                        & 16	& 32	& 64	& 128\\
                \hline
                64	    & 19.48	& 18.33	         & 17.52	      & \textbf{16.93}\\
                128	    & 14.33	& 13.87	         & \textbf{13.09} & 13.70\\
                192	    & 13.39	& 13.31	         & \textbf{12.68} & 13.98\\
                256	    & 11.97	& \textbf{11.56} & 12.65          & 14.33\\
                320	    & 11.46 & \textbf{11.15} & 11.69          & 14.10\\
                384	    & 11.63	& \textbf{11.56} & 12.14	      & 13.75\\
                448	    & \textbf{11.20} & 11.33 & 12.45	      & - \\
                \hline
            \end{tabular}
            }
        \end{table}   

    \subsection{Continuous Speech Recognition}
        Unlike phoneme based ASRs, the proposed AAE based approach does not depend on  linguistic knowledge. It is therefore interesting to compare these approaches on a real-world continues speech recognition (CSR) task. 
        For this purpose, we used the 
        SX records of the TIMIT corpus which contains 450 sentences spoken by 7 speakers, i.e. 3150 utterances in total. We prepared the test set by randomly selecting and putting aside one speaker for each sentence from the SX recordings and used the remaining samples as the training set (450 sentences $\times$ 6 speaker = 2700 utterances). We also included the SA and SI recordings of the TIMIT corpus in the training set.
        The number of AAEs was 384. The number of mixture components in the GMM based phone models was 64. 
        The performance was evaluated in two scenarios: with and without language model. The language model employed in the baseline and the proposed methods is a simple bigram model. 
        
        Table \ref{tab:csr} shows that the proposed AAE-DNN based approach significantly outperforms baseline methods in both scenarios. The performance improvements over the phone based HMM-DNN method in with and without the language model scenarios were 10.68\% and 5.11\%, respectively. 
        The results suggest that the proposed data-driven dictionary and the AAE models are also useful for CSR and a more accurate representation of speech signals can be learned automatically.
        We observed that all 384 AAEs were actually used in the trained dictionary, and the dictionary tend to assign 39\% more HMM states on average to each word as compare with the TIMIT phonetic dictionary. This means that in AAEs, the stay-in-state probability is smaller resulting in more frequent state transitions. This suggests that by using AAEs, the acoustic space was modeled at a higher resolution. This consequently increased the precision of the word pronunciations.

        \begin{table}[t]
            \caption{\label{tab:csr} {\it Comparison of word error rate of each method on continuous speech recognition. In column ”No LM”, no language model was used.}}
            \vspace{2mm}
            \centering{
              \begin{tabular}{ c | c | c } 
                \hline
                Method	    &	No LM	    & 	Bigram \\
                \hline
                Phone GMM	& 	71.11	    & 	43.54 \\
                Phone DNN	&   50.18       & 	20.89 \\
                \hline
                AAE GMM	    & 	59.52	    & 	32.36 \\
                AAE DNN	    & \textbf{39.05} & \textbf{15.78} \\
                \hline
              \end{tabular}
            }
        \end{table}

  \section{Conclusions}
  \label{sec:concl}
    In this work we proposed a novel joint dictionary and sub-word unit learning framework for ASRs. The proposed method does not require linguistic expertise, and can automatically create the set of sub-word units and the corresponding pronunciation dictionary. In our method, reliable pronunciations are estimated from multiple utterances by an efficient approximation of K-dimensional Viterbi algorithm which estimates the most probable HMM state sequence common to multiple utterances of a word. Experimental results show that the proposed method significantly outperforms the phone based methods which even get manually prepared dictionary and hand crafted transcriptions as inputs. We further investigated the effects of the number of data-driven sub-word units and showed that the optimal number of sub-word units is much higher than the total number of HMM states of the 61 phones. 
    The future works will be directed towards applying the proposed method to speech recognition for under-resourced languages and large vocabulary continuous speech recognition tasks.

  \newpage
  \eightpt
  \bibliographystyle{IEEEtran}

  \bibliography{mybib}

\end{document}